\documentclass[letterpaper, 10 pt, conference]{ieeeconf}  %
\UseRawInputEncoding

\IEEEoverridecommandlockouts                              %

\usepackage{amsmath} %
\usepackage{amsfonts} %
\usepackage{amssymb}  %
\usepackage{graphicx}
\usepackage{hyperref}
\usepackage{mathtools}
\usepackage{comment}
\usepackage{algorithmicx}
\usepackage{algpseudocode}
\usepackage[table]{xcolor}

\usepackage{xcolor}
\usepackage{caption}
\usepackage{subcaption}
\usepackage[linesnumbered, ruled,vlined]{algorithm2e}

\usepackage[font=footnotesize,labelfont=footnotesize]{subcaption}
\usepackage[font=footnotesize,labelfont=footnotesize]{caption}

\SetKwInput{KwInput}{Input}                %
\SetKwInput{KwOutput}{Output}              %

\usepackage{amssymb}
\usepackage{amsmath, bm}
\usepackage{verbatim}

\usepackage{multirow}
\usepackage{hhline}

\renewcommand{\vec}[1]{\boldsymbol{\mathbf{#1}}}

\usepackage{cite}

\usepackage{color}

\usepackage[normalem]{ulem}

\usepackage{tabularx}

\DeclareMathOperator*{\argmax}{arg\,max}

\graphicspath{{figures/}}

\definecolor{D}{RGB}{196, 114, 0}
\definecolor{Q}{RGB}{18, 113, 148}

\title{\LARGE \bf
Efficient and Accurate Mapping of Subsurface Anatomy via Online Trajectory Optimization for Robot Assisted Surgery}

\author{Brian Y. Cho and Alan Kuntz%
\thanks{The authors are with the Robotics Center and Kahlert School of Computing,
        University of Utah, Salt Lake City, UT 84112, USA.
        This material is based upon work supported in part by the National Science Foundation under grant number 2133027.
        {\tt\small \{brian.cho,alan.kuntz\}@utah.edu}
        }
}

\usepackage{color}

\begin{document}

\maketitle
\begin{abstract}
Robotic surgical subtask automation has the potential to reduce the per-patient workload of human surgeons. 
There are a variety of surgical subtasks that require geometric information of subsurface anatomy, such as the location of tumors, which necessitates accurate and efficient surgical sensing. 
In this work, we propose an automated sensing method that maps 3D subsurface anatomy to provide such geometric knowledge. 
We model the anatomy via a Bayesian Hilbert map-based probabilistic 3D occupancy map.
Using the 3D occupancy map, we plan sensing paths on the surface of the anatomy via a graph search algorithm, $A^*$ search, with a cost function that enables the trajectories generated to balance between exploration of unsensed regions and refining the existing probabilistic understanding. 
We demonstrate the performance of our proposed method by comparing it against 3 different methods in several anatomical environments including a real-life CT scan dataset. 
The experimental results show that our method efficiently detects relevant subsurface anatomy with shorter trajectories than the comparison methods, and the resulting occupancy map achieves high accuracy.
\end{abstract}

\section{Introduction}
As the global population grows, we are faced with an impending shortage of surgeons, potentially leading to a severe issue in the near future~\cite{Lynge2008_AS}.
It is expected that the shortage of surgeons will increase by $20,000$ over the next two decades~\cite{Satiani2011_JACS}.
This will result in a significant increase in the per-patient workload of surgeons, potentially reducing quality of care in many fields of surgery including general surgery~\cite{Lynge2008_AS,Ellison2020_Surgery}, urology~\cite{McKibben2016_Urology}, and orthopedic surgery~\cite{Ellison2020_Surgery}, among others.  
For example, in cardiothoracic surgery, it is anticipated there will be an $\approx 120\%$ increase of workload per surgeon by $2035$~\cite{Moffatt2018_JTCS}.

One potential approach to alleviating this problem is to automate certain surgical subtasks, such as sensing~\cite{Salman2018_ICRA, Garg2016_CASE, Nichols2015_TRO}, retraction~\cite{Thach2021_ICRA, Patil2010_ICRA, Attanasio2020_RAL}, and cutting~\cite{Wang2022_RAL}.
In addition to reducing per-procedure surgeon workload, such automation may be able to offer increased accuracy and greater dexterity~\cite{Yip2018_EMR}.
Specifically, automating intraoperative sensing using, e.g., a laparoscopic ultrasound probe has the potential to facilitate subsequent surgeon-operated and automated subtasks by providing an accurate geometrical model of subsurface anatomy. 

\begin{figure}[t]
    \centering
    \includegraphics[width=\linewidth]{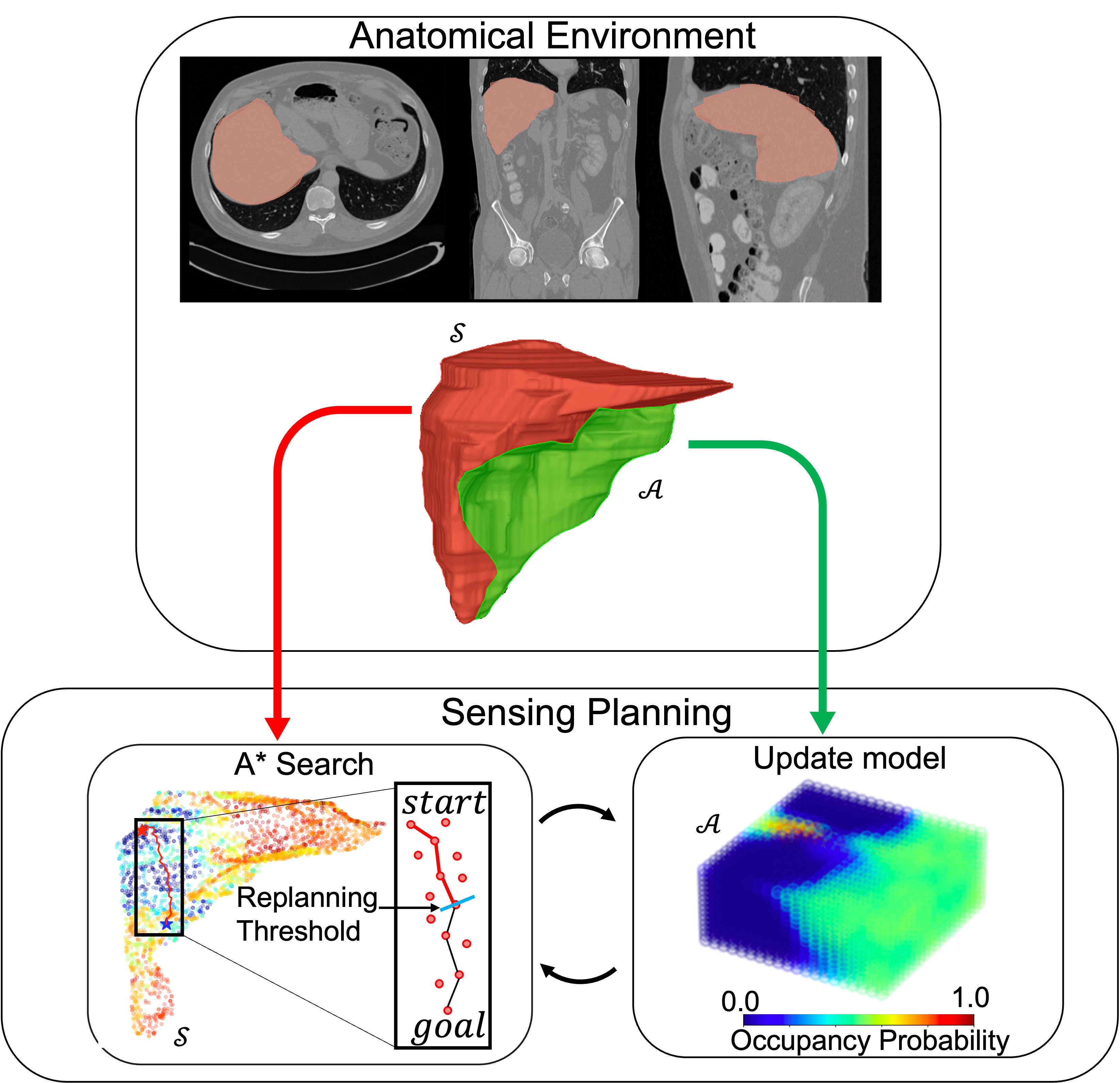}
  \caption{An overview of our proposed method to plan sensing paths on an organ surface. (Top) Given an anatomical surface, such as the surface of a liver during a laparoscopic procedure, we model a 3D probability distribution of the subsurface anatomy as we collect sensor data. (Bottom) We then plan the sensing path on the organ surface (sensor workspace $\mathcal{S}$) based on the 3D occupancy map $\mathcal{A}$ (representing the volume of the organ under the surface) using the graph search algorithm A*. While executing the trajectory, we frequently replan the path to reflect the information gathered during prior portions of the trajectory. This process then repeats until the anatomy is accurately mapped.}
  \label{fig:overview}
\end{figure}

In this work, we propose a method of planning efficient sensing paths to accurately map subsurface anatomy (see Fig~\ref{fig:overview}).
We represent the subsurface anatomy via a Bayesian Hilbert map~\cite{Senanayake2017_PMLR}.
The Bayesian Hilbert map provides a 3D probability distribution over the geometry of anatomical features of interest, such as tumors.
We leverage the map to generate sensing trajectories, e.g., paths to trace on the surface of an organ with a sensor such as an ultrasound probe.
We do so via a graph search algorithm that balances exploration and exploitation, enabling the robot to sense unexplored areas of the anatomy while refining the information about previously sensed regions.

We evaluate our method in multiple simulated surgical scenarios, tasking it with locating subsurface tumors, in both synthetic environments as well as anatomical scenarios segmented from real patient data.
We compare the performance of our proposed method against several alternative methods. 
We further evaluate aspects of our method, such as replanning, and demonstrate their impact on the method's performance.
The results show that our method outperforms three comparison methods, accurately mapping the subsurface tumors in anatomical environments with sensing trajectories that are $20-50\%$ shorter than the comparison methods.

\section{Related Work}
There have been a number of efforts to automate robotic surgical subtasks.
McKinley \textit{et al.}~\cite{McKinley2016_CASE} proposed a system to automate surgical tool changes in tumor resection consisting of palpation, incision, debridement, and injection. 
Thach \textit{et al.}~\cite{Thach2021_ICRA} developed a method of manipulating deformable objects for tissue retraction to expose areas of interest beneath the tissue. 
Elek \textit{et al.}~\cite{Elek2017_INES} developed an automated blunt dissection method. 
To perform the subtasks mentioned above precisely, accurate geometric information of the anatomy is required. 
In this work we focus on automating surgical sensing to provide such information.

Researchers have studied sensing automation through palpation probing.
Nichols \textit{et al.}~\cite{Nichols2015_TRO} proposed an automated palpation method for tumor localization and segmentation.
Garg \textit{et al.}~\cite{Garg2016_CASE} developed a method of localizing tumor boundaries on a stiffness map via palpation.
Salman \textit{et al.}~\cite{Salman2018_ICRA} optimized a palpation trajectory by modeling the stiffness distribution of the anatomical space via Gaussian process regression.
These methods assumed two dimensional cases, specifically.
In this work, we consider 3D cases where the robotic agent uses a 3D sensor (e.g., a swept ultrasound) in a 3D anatomical space and along a non-planar surface.

Effort has also been made to automate ultrasound sensing.
Marahrens \textit{et al.}~\cite{Marahrens2022_Frontiers} developed an autonomous, learning-based ultrasound scanning method that considers the coupling between a sensor probe and the tissue surface for accurate reconstruction of 3D anatomy, focusing on sensor orientation aligning with an area of interest (e.g., a vessel).
Their work, however, considers planning the sensor route specifically around the vessel, which may not be applicable for 3D mapping of more complex anatomical structures. 
Our autonomous sensing approach particularly focuses on generating an efficient sensor path to gather the geometrical information of the 3D anatomy, with a trade-off between exploiting known information and exploring regions of high uncertainty. 
 
The autonomous method must be able to sense regions of interest efficiently and map the anatomical environment accurately.
In our previous work~\cite{Cho2021_ISMR}, we proposed a method to determine discrete sensing sequences by utilizing a Bayesian approach~\cite{Pelikan1999_GECCO} that finds a global extremum of a function (e.g., Bayesian Hilbert maps) through acquisition functions~\cite{Jones1998_JGO}, e.g., expected improvement (EI).
The acquisition function determines where to sample next in a given probability distribution, i.e., in our case the EI value provides guidance on where to sense next to gain more geometric information of subsurface anatomy.
Our prior work was limited to discrete and discontinuous sensing actions, rather than continuous sensing trajectories.
In this work, we leverage these determined sensing sequences, utilizing them as way-points for continuous sensing trajectories.

Path planning via graph search has widely been used for other robotics problems~\cite{Aine2016_IJRR, Koenig2005_TRO, Driess2017_IROS}.
Aine \textit{et al.}~\cite{Aine2016_IJRR} proposed a search approach, called multi-heuristic A*, which leverages multiple inadmissible heuristics and a consistent heuristic while still guaranteeing completeness and bounded sub-optimality.
This mitigates difficulties of designing admissible heuristics and formulating a heuristic that considers all complex cases of the search problems.
Koenig \textit{et al.}~\cite{Koenig2005_TRO} developed an incremental heuristic search method, called D*~Lite.
This has been widely used for sensor-based motion planning for mobile robot navigation, in which it assumes the unknown area of the given environment to be free space (i.e., without obstacles), updating the map as it observes new geometrical information, and replanning the shortest path based on new information. 
In this work, we plan sensing paths via $A^*$ graph search~\cite{Hart1968_TSSC} with a novel edge weight function that combines two different objectives to make the path efficient and informative, i.e., considering information gain along the path itself.

Research has been performed to determine an object's surface shape using a continuous sensing path.
Driess \textit{et al.}~\cite{Driess2017_IROS} developed an active learning method for tactile exploration of an object which improves upon inefficient touch-and-retract exploration strategies~\cite{Dragiev2013_ICRA}.
This method models an object of interest via a Gaussian process implicit surface model that represents the shape uncertainty of the object and guides the exploration path to reduce the measured uncertainty. 
By contrast, in this work we represent the environment of interest via a Bayesian Hilbert map (described in Section~\ref{section:background} and which performs a similar purpose but is parametric) and plan a continuous sensing path to efficiently explore the environment and accurately map 3D subsurface anatomy.

In this work, our goal is to improve the quality of sensing during robot surgery by combining and building upon many of the methods mentioned above.

\section{Background on Bayesian Hilbert maps}
\label{section:background}

A Bayesian Hilbert map is a parametric logistic regression model that leverages kernels to build a continuous occupancy map in dynamic environments.
In this work we utilize Bayesian Hilbert maps~\cite{Senanayake2017_PMLR} to iteratively update an occupancy probability map, i.e., a map probabilistically detailing the location and geometry of a specific anatomical feature in the environment (such as subsurface tumors in an organ).
The Bayesian Hilbert map is composed of $M$ hinge points spatially fixed in the anatomical space which are used as kernel centers.
Kernel functions, $k(\vec{x}, \widetilde{\vec{x}}_{j})$, then evaluate the similarity between any sensed point $\vec{x}$ and each hinge point $\tilde{x}$.
Using the kernel function, we compute feature vectors $\Psi(\vec{x}) = (k(\vec{x}, \widetilde{\vec{x}}_{1}), k(\vec{x}, \widetilde{\vec{x}}_{2}), \cdots)$ to model the occupancy probability of any point using the sigmoid function $\sigma(\cdot)$, i.e.,
\[
    P(y|\vec{x},\vec{w}) = \sigma(\vec{w}\Psi^{T}(\mathbf{\vec{x}})),
\]
where $\vec{w}$ is a weight vector defined as Gaussians $\vec{w}~\sim~\mathcal{N}(\vec{\mu} \in \mathbb{R}^{1 \times M}, \vec{\Sigma} \in \mathbb{R}^{1 \times M})$.
The goal of the Bayesian Hilbert map is then to learn a weight parameter vector $\vec{\omega} = \{ \vec{\mu}, \vec{\Sigma} \}$ as we collect the sensor data.
We then query any point $\vec{x}_{*}$ in a given environment to evaluate the point being occupied in the 3D occupancy map defined as $P(y|\vec{x}_{*},\vec{w})$ (see \cite{Senanayake2017_PMLR} for more details).

\section{Problem Formulation}
\label{section:pdef}

\begin{figure}[t]
    \captionsetup{skip=-0.1pt}
    \centering
    \includegraphics[width=0.9\linewidth]{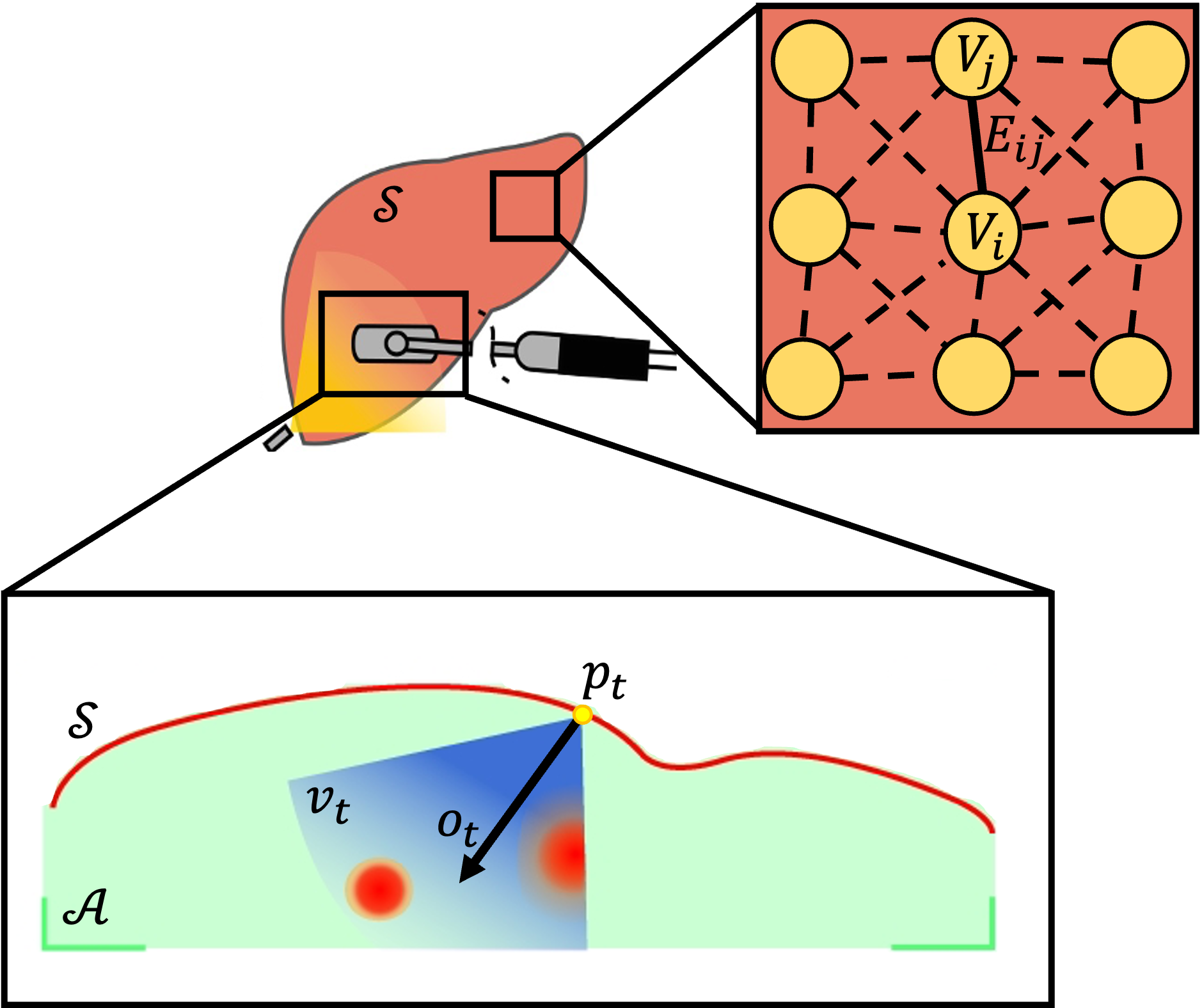}
  \caption{An illustration demonstrating the concept of localizing tumors embedded in an anatomical environment $\mathcal{A}$ under a human organ surface $\mathcal{S}$ with a 3D sensor, e.g., an ultrasound transducer. We define the sensor workspace graph $G_\mathcal{S}$ on $\mathcal{S}$ as a graph composed of vertices $V_i$ and edges $E_{ij}$. The surgical agent then moves along the graph with a cone-shaped volumetric sensor which at any instant in time can be defined by its sensing position $\vec{p_t}$ and orientation $\vec{o_t}$, and a corresponding sensor volume $v_t$. 
  }
  \label{fig:prob}
\end{figure}

The goal of this work is to plan continuous sensing paths to map an anatomical environment of interest.
In this work we adopt assumptions similar to those used in~\cite{Cho2021_ISMR}, i.e., that the anatomy is rigid and that a sensor is noiseless.
    
We consider the case where a surgical agent uses a 3D sensor on a sensor workspace $\mathcal{S}$ consisting of $N$ possible sensing poses $\vec{s}_{i}$ for $i = {1, \cdots, N}$, where $\mathcal{S}$ is, e.g., the 2D surface of an organ.
The goal is to improve the geometrical understanding of the anatomical environment $\mathcal{A} \subset \mathbb{R}^{3}$ beneath $\mathcal{S}$ in which some anatomical features of interest $\mathcal{T} \subset \mathcal{A}$, e.g., tumors, are embedded. 
We define a sensing pose $\vec{s}_{i}$ as a vector $\vec{s}_{i} = [\vec{p}_{i}, \vec{o}_{i}]$, where $\vec{p}_{i} \in \mathbb{R}^{3}$ and $\vec{o}_{i} \in SO(3)$ are the sensing position and orientation, respectively (where $\vec{p}_{i}$ is on $\mathcal{S}$ and $\vec{o}_{i}$ points into $\mathcal{A}$ from $\mathcal{S}$).

We next define a graph on the sensor workspace $\mathcal{S}$ composed of a set of vertices and edges $G_\mathcal{S} =$ ($V$, $E$).
Each vertex $V_i$ corresponds to a sensing pose $\vec{s}_{i}$ on $\mathcal{S}$.
We connect each vertex with an edge in the graph to its 8 adjacent vertices (see Fig.~\ref{fig:prob}).

We define a sensor model (e.g., as an abstraction of an ultrasound transducer) as a cone-shaped volume $v_i \subset \mathbb{R}^3$ with pose corresponding to a sensing pose $\vec{s}_{i}$.
The sensor volume $v_i$ contains data points $\vec{x}_i \in v_i$.
For any sensing action $t$, we collect sensor data $\mathcal{D}_t$ consisting of a set of tuples $(\vec{x}^t, y^t)$ where an occupancy label $y \in \{0, 1\}$ indicates whether the point $\vec{x}$ is part of the anatomical feature being mapped, i.e., a point in the cone $\vec{x} \in v_t$ can be labeled as either occupied ($y = 1$) or unoccupied ($y = 0$) by the anatomy type of interest (e.g., tumor). 
We then use the sensor data $\mathcal{D}_t$ to update the Bayesian Hilbert map parameters that define an occupancy probability of the anatomy.

We define a sensing trajectory $r$ as a path that traverses the graph, passing through an ordered set of $k$ vertices, i.e., $\mathcal{S}_{\mathcal{M}}^r  = \{V_{1}, \cdots, V_{t}, \cdots, V_{k} \} \subset \mathcal{S}$ where a sensing action is performed at finely discretized locations along the edges and at the vertices in the path, which is used to update the Bayesian Hilbert map. 
We then define the sensed volume of the path at iteration $r$, 
\[
\mathcal{V}_{\mathcal{S}_{\mathcal{M}}^r} = 
\bigcup_{i=1}^{k} v_{i}.
\]
We define the length of a trajectory to be the sum of Euclidean distances between all adjacent vertices in the path and leverage this length as a metric of efficiency.

The problem then becomes to iteratively plan a sequence of sensing trajectories that maps the anatomy with high accuracy, defined as an ability to correctly classify positive samples (i.e., tumors), while minimizing the total path length.

\section{Method}

\begin{algorithm}[t]
\SetAlgoLined
\KwInput{Sensor workspace $\mathcal{S}$, Search space $\mathcal{A}$, \\ Graph $G_\mathcal{S} =$ ($V$, $E$)}
 Initialize the BHM kernels $\vec{\omega} \leftarrow \vec{\mu}_{0}, \vec{\Sigma}_{0}$ \\ 
  Trajectory iteration $r=0$ \\
  $\vec{s}_{0}$ $\leftarrow$ an $s$ sampled at random from $\mathcal{S}$ \\
  \While{termination criteria not met}{
  $r = r + 1$\\
  \eIf{r=1}{
  Collect sensor data $\mathcal{D}^{1}$ on $\vec{s}_{0}$ \\ 
  }
  {
  Collect sensor data $\mathcal{D}^{r}$ along a path $\mathcal{S}_{\mathcal{M}}^r$\\ 
  }
  $\vec{\mu}_{r}, \vec{\Sigma}_{r} \leftarrow$ \texttt{learn$\_$params}($\mathcal{D}^{r}$, $\vec{\omega}$) \Comment{Sec.~\ref{section:background}}\\
  $\vec{\omega} \leftarrow \vec{\mu}_{r}, \vec{\Sigma}_{r}$ \Comment{Sec.~\ref{section:background}}\\
  $EI \leftarrow$ \texttt{update$\_$EI}($\vec{x_*}, \vec{\omega}$) \Comment{Eq.~\ref{eq:ei}}\\
  $g(V_i)$ = \texttt{compute$\_$vertex$\_$value}($EI$) \Comment{Eq.~\ref{eq:vertex_value}}\\
  $V_{k}$ = \texttt{set$\_$goal$\_$vertex}($g(V_i)$) \Comment{Sec.~\ref{section:path_planning}}\\
  $E_{ij}$ = \texttt{update$\_$edge$\_$cost}($V$, $E$, $EI$) \Comment{Eq.~\ref{eq:edge_cost}}\\
  $\mathcal{S}_{\mathcal{M}}^r$ = \texttt{graph$\_$search}($V$, $E$) \Comment{Sec.~\ref{section:path_planning}}\\
  (Optional)\texttt{path$\_$replanning}($\mathcal{S}_{\mathcal{M}}^r$) \Comment{Sec.~\ref{section:path_planning}}\\
 }
 \caption{Sensing Path Planning}
 \label{Alg: path_planner}
\end{algorithm}

This section describes our method for planning and replanning of sensing paths that, based on the occupancy map, obtains as much information as possible during the execution of the paths rather than just choosing the shortest path.
In this way, the method efficiently builds an accurate 3D map of the sub-surface anatomy.

\subsection{Method Overview}
\label{section:method_overview}
This section presents a high-level overview of our proposed method, described in Algorithm~\ref{Alg: path_planner}. 
We begin by initializing the probability distribution of the Bayesian Hilbert maps with zero mean and high variance (line $1$), implying that we have no prior information of the patient's sub-surface anatomy.
We first place the sensor at a sensing pose $\vec{s}_{0}$, chosen uniformly at random from $\mathcal{S}$ (line $3$).
We then collect sensor data (lines $6-10$).
Based upon the sensor data we update the posterior distribution of the occupancy probability (lines $11-12$, see Section~\ref{section:background} and \cite{Senanayake2017_PMLR}) that is then used to update the EI values on the sensor workspace (lines $13-14$, see Section~\ref{section:sensor_workspace}).
We then choose a first goal point that has the highest EI value on the sensor workspace $\mathcal{S}$, meaning the highest probability of attaining as much geometrical information as possible, for path planning from $\vec{s}_{0}$ (line $15$).
With the updated EI values, we calculate the edge costs on $G_\mathcal{S}$ (line $16$) and plan the sensing path using the $A^*$ search algorithm (line $17$, see Section~\ref{section:path_planning}).
We collect data along a portion of the planned path, up to a replanning threshold (line $18$).
The pose at which the sensor stopped is then used as a new start pose, a new goal is chosen, and the algorithm repeats the planning/execution process.

\subsection{Graph Representation of the Sensor Workspace}
\label{section:sensor_workspace}
\begin{figure}[t!]
    \captionsetup{skip=-0.2pt}
    \centering
    \includegraphics[width=\linewidth]{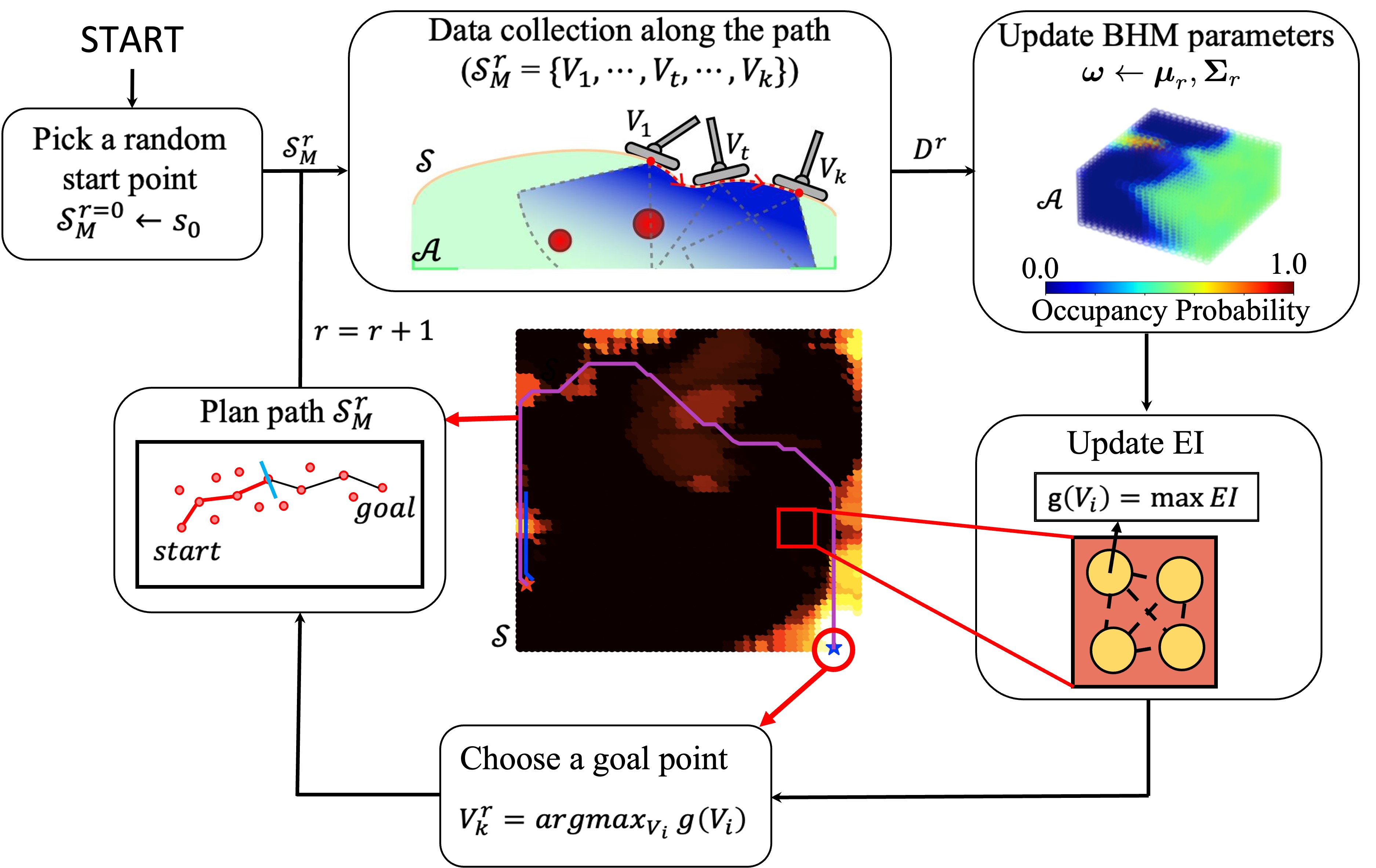}
  \caption{An overview of our proposed method, described in Algorithm~\ref{Alg: path_planner}. We start by choosing the first sensing position on the sensor workspace. We then collect data along the sensing path (it collects data on the first sensing position at the first iteration of the process), update the Bayesian Hilbert map (BHM) parameters in the search space $\mathcal{A}$, and update the vertex values on the sensor workspace $\mathcal{S}$. We then determine a goal point, followed by a path planning via $A^*$ search. The middle figure above shows an example scene of a sensing path on the sensor workspace planned by our method. A brighter color indicates it is more likely to gain geometrical information of embedded anatomy. The purple line shows the path without replanning while the blue shows the one with replanning. Our method plans the path through the areas that are more likely to detect the underlying anatomy. Replanning enables our method to plan paths with up-to-date information.} 
  \label{fig:replanning}
\end{figure}
Based on the Bayesian Hilbert map at each iteration (the while loop, line 4 of Algorithm~\ref{Alg: path_planner}), we first must determine a sensing path over the sensor workspace $\mathcal{S}$, the anatomical surface on which the sensor moves.
It is crucial for a path to balance between exploration and exploitation as insufficient exploration may lead to leaving areas where critical anatomy is unsensed while inadequate exploitation may result in not fully mapping partially identified anatomy.
To do so, we leverage Expected Improvement (EI), a widely used acquisition function in Bayesian optimization.
More formally, the EI (\texttt{update$\_$EI} in Algorithm~\ref{Alg: path_planner}, line 13) is defined in a closed form for any query point $\vec{x}_{*} \in \mathcal{A}$ as:
\begin{equation}
    EI(\vec{x}_{*}, \vec{\omega}) = (\vec{\mu}(\vec{x}_{*}) - f(\vec{x}^{+}) - \xi)\Phi(Z) + \vec{\Sigma}(\vec{x}_{*}) \phi(Z)
    \label{eq:ei}
\end{equation}
where 
\[Z = \frac{\vec{\mu}(\vec{x}_{*}) - f(\vec{x}^{+}) - \xi}{\vec{\Sigma}(\vec{x}_{*})},\] 
$f(\vec{x}^{+})$ is the current maximum occupancy likelihood, $\Phi(\cdot)$ is the cumulative distribution function, $\phi(\cdot)$ is the probability density function, and $\xi$ is an exploration parameter.

We then leverage the sensor workspace graph $G_\mathcal{S}$ (defined above in Section~\ref{section:pdef}).
We augment each vertex with the maximum expected improvement value of points $\vec{x}_i$ within its corresponding sensor volume $v_i$ obtained when we sense at that vertex $V_i$ (\texttt{compute$\_$vertex$\_$value} in Algorithm~\ref{Alg: path_planner}, line 14), i.e., the value of a vertex is defined as
\begin{equation}
    g(V_i) = \max_{\vec{x}_i \in v_i} EI(\vec{x}_i, \vec{\omega}).
    \label{eq:vertex_value}
\end{equation}
Each edge represents the sensing action between vertices and has its own cost of performing a corresponding sensing action along the edge.
We define the edge cost as the combination of two sub-costs (\texttt{update$\_$edge$\_$cost} in Algorithm~\ref{Alg: path_planner}, line 16):
\begin{equation}
  U = U_{dist} + \lambda U_{EI}  
    \label{eq:edge_cost}
\end{equation}
where $U_{dist}$ is the Euclidean distance between the two vertices' locations, $U_{EI}$ is an expected improvement cost, and $\lambda > 0$ is a scaling factor.
Specifically, we define $U_{EI}$ as the reciprocal of the line integral: 
\[ U_{EI} = \frac{1}{\int_{E_{ij}} EI dE_{ij}}. \]
With values at vertices and costs defined over edges we can then leverage graph search algorithms, such as A* search, to plan sensing paths on the sensor workspace by searching over $G_\mathcal{S}$.

\subsection{Sensing Path Planning} 
\label{section:path_planning}

With the graph $G_\mathcal{S}$ defined on the sensor workspace $\mathcal{S}$, we then utilize A* graph search (\texttt{graph$\_$search} in Algorithm~\ref{Alg: path_planner}, line 17), an informed search algorithm proven to be complete and optimal, to efficiently obtain useful information during the trajectory execution. 
We compute the heuristic $h$ for A* as the Euclidean distance from the current vertex $V_t$ to a goal vertex $V_k$ determined as the vertex with the largest expected improvement, i.e., $V_{k} = \argmax_{V_i} g(V_i)$ (\texttt{set$\_$goal$\_$vertex} in Algorithm~\ref{Alg: path_planner}, line 15).
Note that in our previous work~\cite{Cho2021_ISMR}, the goal vertex was utilized for discrete sensing sequences; however, in this work, it serves as the way-points for continuous sensing trajectories.
The A* path is optimal as for all vertices the heuristic function $h(V_t)$ is admissible by its definition, i.e., $0 \leq h(V_t) \leq h^*(V_t)$ where $h^*$ is the true optimal forward cost. 
We can prove $h(V_t) \leq h^*(V_t) = \sum^k_{j=t} U_j $ where $U = U_{dist} + \lambda U_{EI}$ by noting that $\lambda U_{EI}$ is always positive and $U_{dist}$ from the current vertex to the goal is always greater than or equal to the heuristic function $h(V_t)$.

In this way, we are able to plan the sensing path at each iteration given the Bayesian Hilbert map.
However, we must update the map after short sensing sequences to leverage the information gathered during sensing.
We thus only execute a short portion of the path and then replan (\texttt{path$\_$replanning} in Algorithm~\ref{Alg: path_planner}, line 18).
We do so by ensuring that each iteration ends when the path reaches a threshold length (see Fig.~\ref{fig:replanning}). 

\section{Experiments and Results}

To evaluate our method we consider the case of identifying and mapping subsurface tumors embedded in an organ.
We demonstrate experimental results for our method in two different environment types, synthetic scenarios and real medical scenarios.
We generate the synthetic scenarios via a curved surface and place multiple tumors with random locations and of random sizes below the surface.
For the medical scenarios, we segment liver surfaces and liver tumors from real patient CT scans in the liver tumor segmentation (LiTS) dataset~\cite{bilic2023liver}, comprised of $201$ CT images of the abdomen. We segment the liver and tumors using 3D Slicer~\cite{3DSlicer}.

We evaluate our method's efficiency and accuracy against 3 different comparison methods, a straight line planner, $A^*$ with only distance as edge weights, and Dijkstra's algorithm with only $EI$ as edge weights.
We also compare our method to a version without replanning, to evaluate the improvement when using replanning.

For all environments and experiments, we use a cost scaling factor $\lambda$ of $70$ and a replanning threshold of $6$ cm, determined by grid search on one of the LiTS scenarios.
We also use the exploration parameter $\xi = 0.01$ for the expected improvement function and $3468$ equally spaced hinge points for the Bayesian Hilbert map, which were set experimentally.

\subsection{Evaluating Efficiency}
\label{section:efficiency_evaluation}
We evaluate a method's efficiency by measuring the total arc length of the sensing path generated by the method, defined as \[|\mathcal{S}_{\mathcal{M}}| = \sum_{r}|\mathcal{S}^r_{\mathcal{M}}|. \] 
We measure the total arc length of each method's path required to sense 95\% of the embedded tumor geometries.
For evaluation in all environments we discretize each tumor with 500 data points and discretize the organ surface (the sensor workspace) with 3600 vertices.
For all scenarios we precompute the orientation of each sensing pose as the surface normal perpendicular to a tangent plane that fits the 10 nearest neighbors of each surface point.
\begin{figure}[ht]
    \captionsetup{skip=-1pt}
    \centering
    \includegraphics[width=\linewidth]{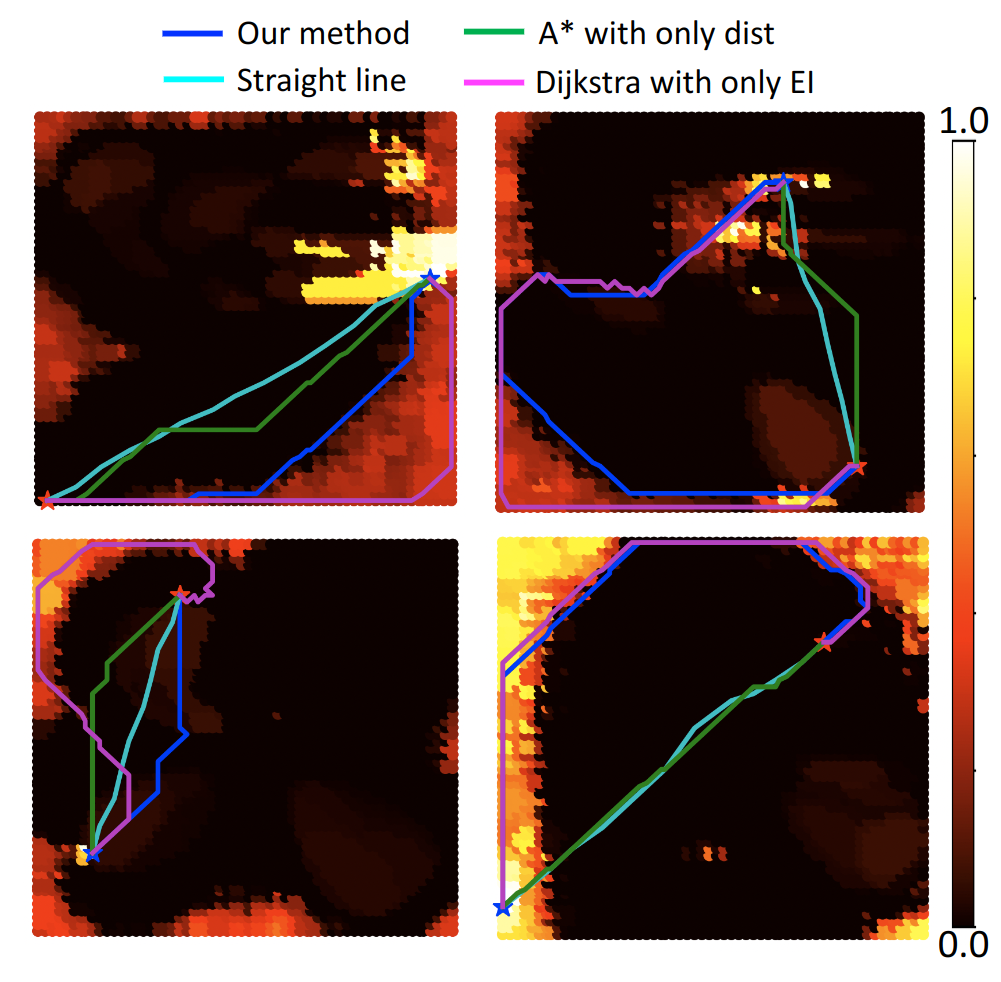}
  \caption{Four example scenarios with the planned paths generated by our method and the comparison methods from one start to one goal (one iteration of planning), blue: our method, cyan: straight line planner, green: $A^*$ with only distance cost, and purple: Dijkstra's algorithm with only EI cost. Bright color in the heat map represents areas with high EI and dark colors areas with low EI. Our method plans a path passing through areas where it will collect more information while still balancing this objective with path length. The straight line planner chooses a path going straight toward the goal point. $A^*$ with only distance cost plans the shortest path on the surface. Dijkstra's algorithm with only EI cost plans a path that maximizes EI but does not account for path length.}
  \label{fig:comparison_methods}
\end{figure}

\begin{figure}[t]
    \captionsetup{skip=-0.5pt}
    \centering
   \begin{subfigure}[b]{0.5\textwidth}
   \captionsetup{skip=-0.5pt}
	\includegraphics[width=\textwidth]{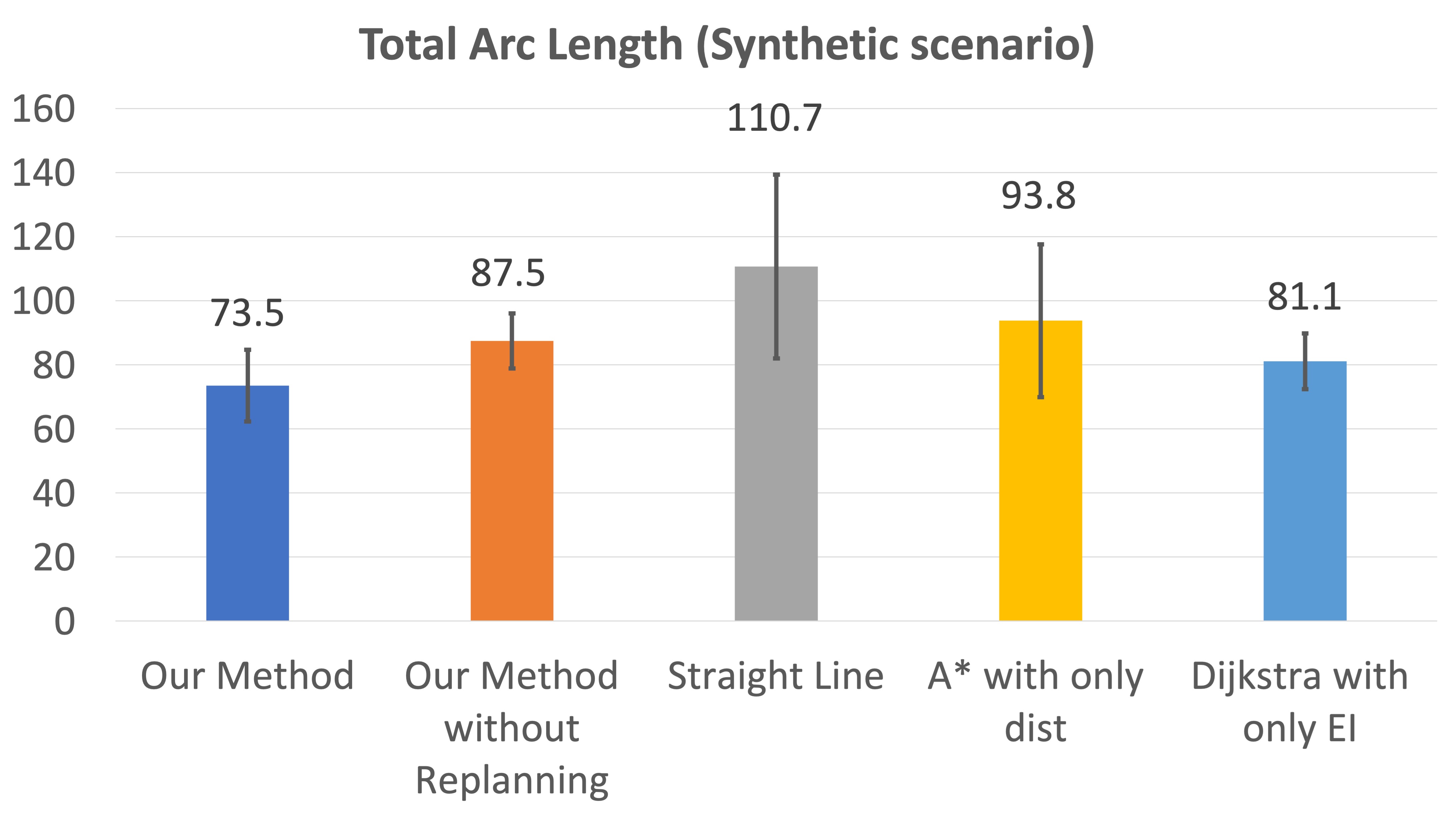}
    \caption{Synthetic scenario}
    \label{fig:comp_Synthetic}
  \end{subfigure}
  \begin{subfigure}[b]{0.5\textwidth}
  \captionsetup{skip=-0.5pt}
	\includegraphics[width=\textwidth]{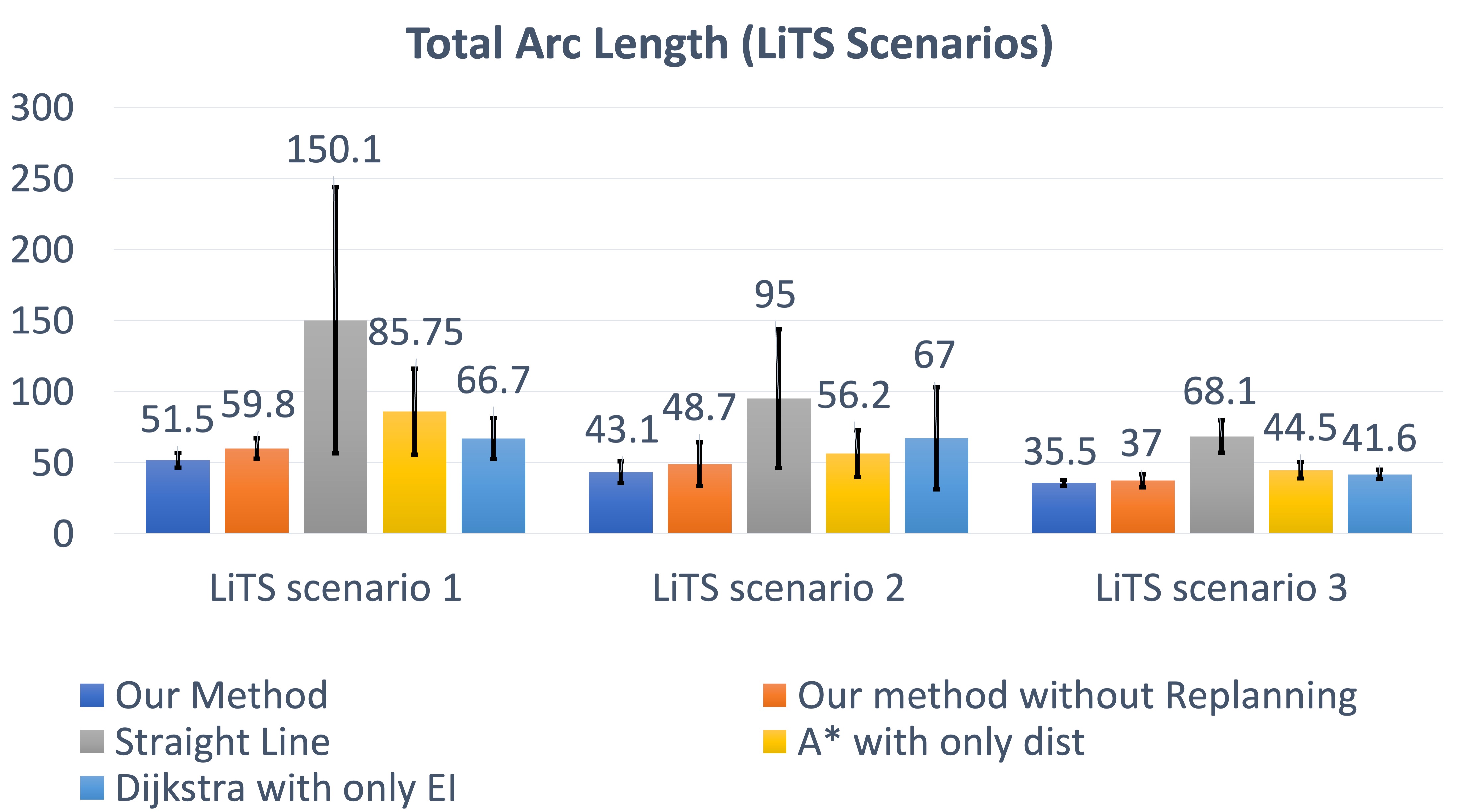}
    \caption{LiTS scenario}
    \label{fig:comp_LiTS}
  \end{subfigure}
  \caption{Efficiency comparison with respect to total arc length (y-axis) taken to detect 95\% of all tumor geometries. All units are in cm. (a) For all synthetic scenarios, the mean and standard deviation are $73.5 \pm 11.2$ for our method, $87.5 \pm 8.6$ for our method without replanning, $110.7 \pm 28.7$ for a straight line planner, $93.8 \pm 23.9$ for $A^*$ with only distance cost, and $81.1 \pm 8.7$ for Dijkstra's with only EI cost. (b) For the three LiTS scenarios, the results show $43.4 \pm 5$, $48.5 \pm 9.1$, $104.4 \pm 51.3$, $62.2 \pm 17.5$, and $58.4 \pm 18$ on average for the respective methods.}
  \label{fig:eff_comparison}
\end{figure}

In Fig.~\ref{fig:comparison_methods} we show qualitative results for our method and the comparison methods.
Each subfigure in Fig.~\ref{fig:comparison_methods} presents a top-down view of the sensor workspace with color indicating EI value during an intermediate iteration of sensing planning on a synthetic scenario, with the red and blue stars denoting the start and goal vertices, respectively.
For the straight line planner (SL), we draw a straight line in 3D from the current point to the goal point and project the line onto the sensor workspace (passing through all vertices in the sensor workspace placed along the line). 
$A^*$ with only distance cost (AD) searches for the goal and generates a sensing path using $A^*$ with the same heuristic function as our method but with an edge cost composed only of $U_{dist}$.
For Dijkstra's algorithm with only EI cost (DE), we use the edge cost function composed only of the EI cost $U_{EI}$.

We perform 100 trials in the synthetic scenario in which we evaluate 10 random sizes and locations of the embedded tumors, each for 10 random initial sensing positions.
For the LiTS scenarios, we experiment on 3 different patient scans randomly chosen from the $201$ CT images.
We segment these scenarios and perform $10$ trials for each scenario with random start positions, averaging the results.  
Compared to our method for all scenarios, the comparison methods (SL, AD, DE) require 2.1, 1.4, and 1.26 times longer total arc length on average, respectively to sense the embedded tumors (see Fig~\ref{fig:eff_comparison}).
For all scenarios, these results show that our method outperforms the other methods in the arc-length required to detect the tumors, a proxy for efficiency.

We also evaluate our method with and without replanning, characterizing the benefit of the algorithm's ability to update the EI map as it senses.
We find that replanning provides 14\% improvements on average in path length (see Fig~\ref{fig:eff_comparison}).

\subsection{Evaluating Accuracy}
In addition to efficiency, the goal of this work is to plan a sensing path on the organ surface to map the subsurface anatomy accurately.
We thus evaluate the accuracy of the resulting occupancy map generated by our method for both the synthetic scenarios and the LiTS scenarios.

We note that our method generates a 3D probabilistic occupancy map that acts as a binary classifier to determine whether any given point in the environment is occupied by the target anatomy type, e.g., tumor, or not.
Two popular metrics for evaluating the accuracy of a binary classifier are the area under the receiver operating characteristics (AUROC) and the area under the precision-recall curve (AUPRC).
When evaluating classification of a highly imbalanced dataset (i.e., a set in which there is a large difference between the number of positive and negative data points), AUROC may give an overly-optimistic view of the classifier's performance potentially leading to incorrect interpretations~\cite{Saito2015_PLOS}. 
AUPRC by contrast is known to be suitable for highly imbalanced datasets.
As our dataset is imbalanced (with a 300:1 ratio of free space to occupied space), as in~\cite{Cho2021_ISMR} we choose AUPRC as our metric for evaluating accuracy.

\begin{figure}[t!]
    \captionsetup{skip=-0.5pt}
    \centering
    \includegraphics[width=0.98\linewidth]{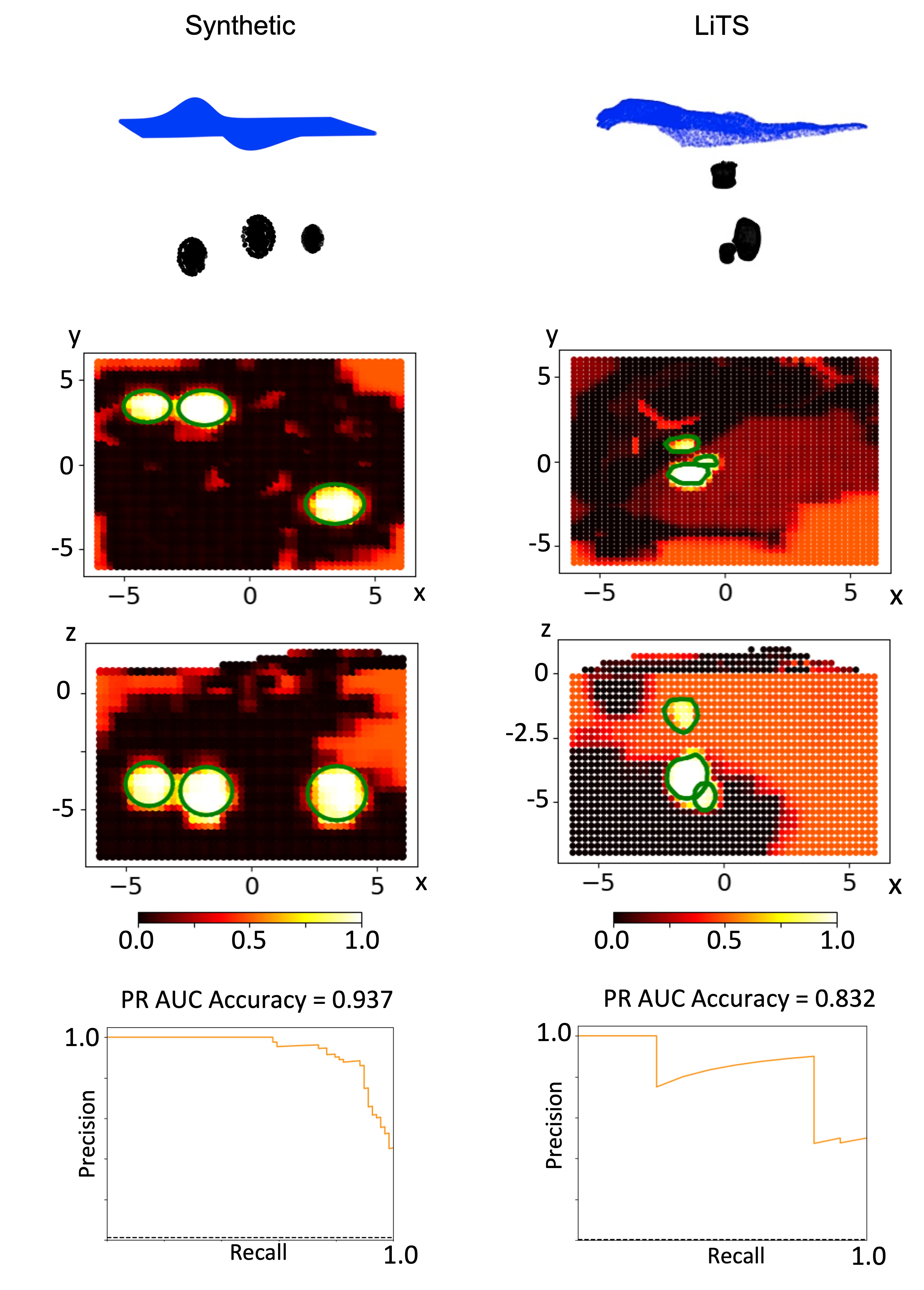}
  \caption{Accuracy evaluation on the final occupancy map of the synthetic scenario for the $100^{\mathrm{th}}$ trial (left column) and the $3^{\mathrm{rd}}$ LiTS scenario for the $10^{\mathrm{th}}$ trial (right column). The heat map represents the occupancy probability of the anatomy projected onto each plane. (Top) The 3D visualization of the environment and tumor ground truth. ($2^{\mathrm{nd}}$ row) The occupancy map is projected onto the x-y plane. Green is the true location of the projected tumors. ($3^{\mathrm{rd}}$ row) The occupancy map is projected onto the x-z plane (side view). (Bottom) The AUPRC is about $0.937$ in the synthetic scenario and $0.832$ in LiTS scenario, respectively.} 
  \label{fig:acc_eval_synthetic}
\end{figure}

As an example, Fig~\ref{fig:acc_eval_synthetic} (left) shows both a top view and a side view of the final-state occupancy map at the last iteration for the 100th trial of the synthetic scenario, including the ground-truth tumor location and the corresponding AUPRC graph. 
In the synthetic scenario, the occupancy map matches the true tumor location with a high AUPRC of 0.937 (noting that a random classifier would display an AUPRC of 0.03).
The average AUPRC for all synthetic scenarios (i.e., averaged over the 100 trials) is $0.903 \pm 0.028$.
Fig~\ref{fig:acc_eval_synthetic} (right) shows the resulting occupancy map at the last iteration of the 10th trial for the third LiTS scenario. 
The LiTS scenario also shows high accuracy with an AUPRC of 0.832, considering that a baseline classifier would have an AUPRC of 0.003.
The AUPRC on average for all LiTS scenarios (i.e., averaged over the 30 trials) is $0.814 \pm 0.009$.
Compared to the performance of the random classifiers in both scenarios, the occupancy maps generated by our method are highly accurate. 

\section{Conclusion}

In this work, we propose a method that plans an informative sensing path on an organ surface to localize subsurface anatomy both efficiently and accurately.
Our method leverages the $A^*$ search algorithm with an edge-weight function that enables the method to balance exploration and exploitation using a Bayesian Hilbert map representation of the anatomical environment.
We evaluate both the efficiency and accuracy of our method by using total arc length and AUPRC as metrics.
We compare our method against 3 different methods, a straight line planner, $A^*$ with only distance cost, and Dijkstra's algorithm with only EI cost. 
On average our method significantly outperforms all comparison methods for all scenarios, in both synthetic environments and real-life tumors segmented from patient CT scans.

In future work we plan to apply our proposed method to real life sensing, e.g., an ultrasound probe on a physical surgical system, e.g., the da Vinci Research Kit surgical robot~\cite{Kazanzides2014_ICRA_DVRK}.
We also plan to augment our method to consider tissue properties, e.g., deformation caused by a sensing action, as well as prior information on likely subsurface anatomical mappings based on pre-operative sensing.

\bibliographystyle{IEEEtran}

\bibliography{reference}

\end{document}